\documentclass[10pt,twocolumn,letterpaper]{article}

\usepackage{cvpr}
\usepackage{times}
\usepackage{epsfig}
\usepackage{graphicx}
\usepackage{amsmath}
\usepackage{amssymb}

\usepackage[bookmarks=false,colorlinks=true,linkcolor=black,citecolor=black,filecolor=black,urlcolor=black]{hyperref}

\cvprfinalcopy 

\def\cvprPaperID{5540} 

\ifcvprfinal\fi
\begin{document}

\title{Residual Convolutional Neural Network Revisited with Active Weighted Mapping}

\author{HyoungHo Jung\\
Dept. of Computer Engineering, Ajou University\\
\and
Ryong Lee\\
KISTI\\
\and
Sanghwan Lee\\
KISTI\\
\and
Wonjun Hwang\\
Dept. of Computer Engineering, Ajou University\\
{\tt\small \{hhjung1202, wjhwang\}@ajou.ac.kr}
}

\maketitle

\begin{abstract}
In visual recognition, the key to the performance improvement of ResNet is the success in establishing the stack of deep sequential convolutional layers using identical mapping by a shortcut connection. It results in multiple paths of data flow under a network and the paths are merged with the equal weights. However, it is questionable whether it is correct to use the fixed and predefined weights at the mapping units of all paths. In this paper, we introduce the active weighted mapping method which infers proper weight values based on the characteristic of input data on the fly. The weight values of each mapping unit are not fixed but changed as the input image is changed, and the most proper weight values for each mapping unit are derived according to the input image. For this purpose, channel-wise information is embedded from both the shortcut connection and convolutional block, and then the fully connected layers are used to estimate the weight values for the mapping units. We train the backbone network and the proposed module alternately for a more stable learning of the proposed method. Results of the extensive experiments show that the proposed method works successfully on the various backbone architectures from ResNet to DenseNet. We also verify the superiority and generality of the proposed method on various datasets in comparison with the baseline.
\end{abstract}

\section{Introduction}
\begin{figure}
    \begin{center}
    \includegraphics[width=8.0cm]{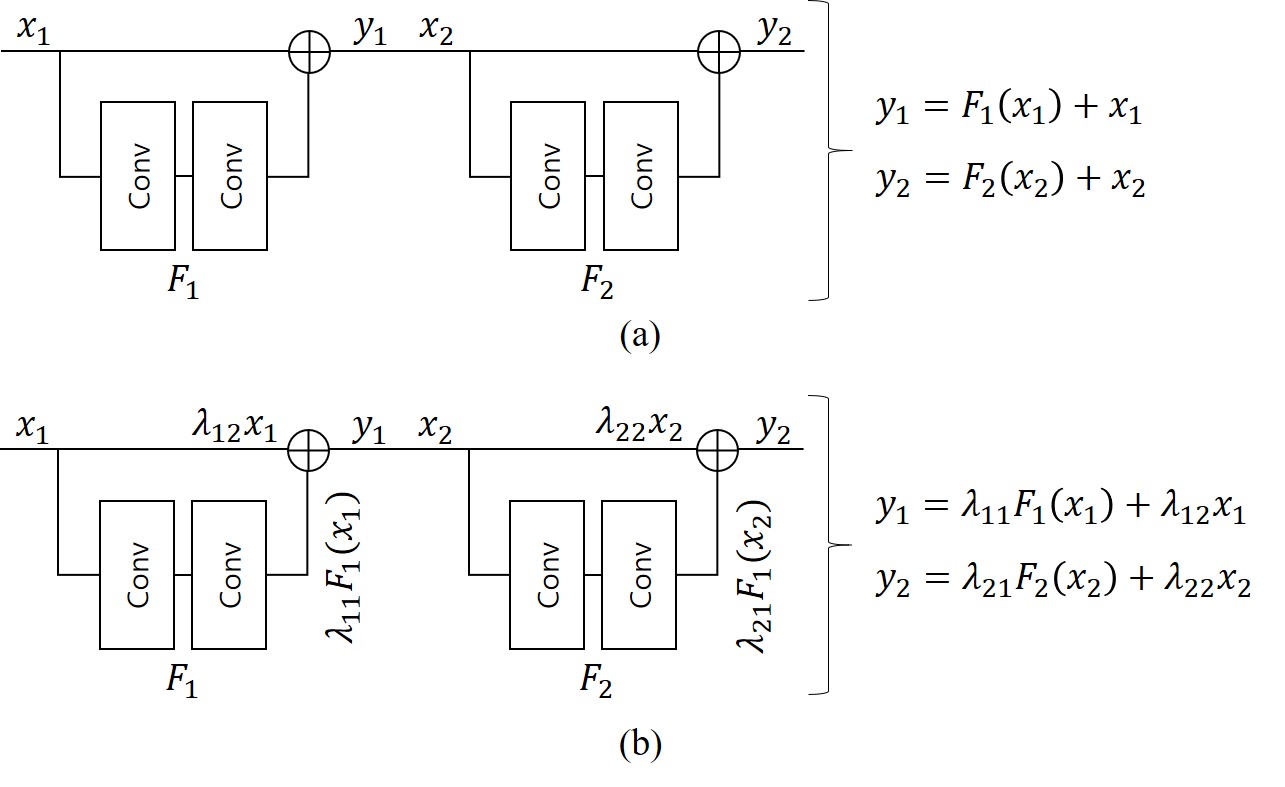}
    \end{center}
    \caption{Schematic comparison between (a) Identical mapping at ResNet~\cite{Res}, and (b) the proposed active weighted mapping.}
    \label{fig:1}
\end{figure}

It has recently been noted that deeper stacking of the layers of a convolutional neural network lead to better accuracy of the visual recognition. A key challenge in visual recognition has been how to stack a larger number of layers efficiently. Several studies~\cite{Ale}\cite{Hig}\cite{VGG}\cite{Goo}\cite{Res} have been done for this purpose. ResNet~\cite{Res}, which adds shortcut connections to implement identity mapping, offers a simple and effective method to more deeply stack convolutional layers without the gradient vanishing problem. After ResNet~\cite{Res}, many novel trials~\cite{Wid}\cite{Sto}\cite{Den}\cite{Pyr} have focused on determining methods to develop efficient network architectures to ensure better accuracy in visual recognition. However, most of these methods have modified only the main architecture of the neural network based on the identical mapping using the shortcut connection suggested by ResNet~\cite{Res} and achieved high performances in visual recognition. For example, WideNet~\cite{Wid} involved a wide residual network architecture in which the sixteen layer-based wide network outperformed the one hundred layer-based ResNet~\cite{Res}. DenseNet~\cite{Den} was introduced for connecting each layer to every other layer in a feed-forward fashion, and it achieved the best result in the recent ImageNet competition. PyramidNet~\cite{Pyr} proposed the residual style-based CNN network based on gradually increased feature map dimensions.

From another viewpoint, Veit et al.~\cite{Ens} considered the shortcut connections of the ResNet~\cite{Res} and suggested that ResNet could be considered to behave as an ensemble of relatively shallow networks. For supporting this argument, the authors removed the residual units individually and proved that significant performance degradation did not occur. Going one step further, Veit et al.~\cite{AIG} recently proposed ConvNet that uses adaptive inference graphs (AIG) for deciding the minimum required layers required for each input image on the fly; the proposed method adaptively defined their network topology conditioned on the input image. In the paper by Squeeze and excitation (SE)~\cite{SE}, simply stacking many series of convolutional layers did not help improve the accuracy of the image classification. To improve the representational power of a network using the modelling the interdependencies between the channels of its convolutional layers, the feature re-calibration method was introduced.

In this paper, we propose the use of active weighted mapping (AWM) method depending on an input image instead of identity mapping by shortcuts, as done by ResNet~\cite{Res}. During residual learning at ResNet~\cite{Res}, it was suggested that identity mapping is enough for visual recognition owing to its simplicity. However, our basic assumption is that although identical mapping helped improve the accuracy of deep learning methods early, it is now a barrier to further performance improvements. To overcome this challenge, we modify the identical mapping to the active weighted mapping using a short connection under the assumption that the mapping modules of the different layers need different weight values for better accuracy. As compared in Fig.~\ref{fig:1}, the main difference between the identical mapping and the proposed method is that the proposed active weighted mapping method provides the proper weighting factors for each mapping layer according to the identity of the input image on the fly, whereas the identical mapping of ResNet~\cite{Res} simply assumed that two paths are merged by the sum operation. Moreover, we qualitatively prove that the weight value of each mapping layer is different according to the class of the input image. For evaluating the importance of a ConvNet unit $F(\mathbf{x})$ or shortcut connection $\mathbf{x}$, we extract the global convolutional information embedded using the global average pooling layer of each unit. The global embedding features are concatenated, and the dimensionality is suitably reduced by fully connected (FC) layers. Subsequently, we use the sigmoid function to determine the different weights. For the stable end-to-end learning of the proposed method, we learn the backbone architecture and the AWM unit alternately in the training stage. Note that when one side is learning, the other side is frozen and there is no parameter update. In this respect, we can model the different weighted mapping units of the deep residual network and validate that these AWM units generate the different weights according to the class of the input image. Moreover, as an ablation test, we train the linear discriminant analysis (LDA)~\cite{LDA} model using the features obtained from the different weight values of each mapping layer at the deep learning stage and confirm that the inferred weight features fundamentally have the power to classify visual objects despite their small dimensionality.

The three main contributions of this paper are as follow: (1) The active weighted mapping method is proposed to improve the basic performance of ResNet, in which the weight values are changed according to the class of an input image. (2) By using the ablation test using the LDA method, we prove that the inferred weight values can be used to classify the visual objects. (3) The proposed method is validated by performing extensive experiments using the basic ResNet~\cite{Res} to DenseNet~\cite{Den} with Cifar-10, Cifar-100, and ImageNet 2012 datasets.

The remainder of this paper is organized as follows. Section 2 presents the related works and highlights the novelty of the proposed method. Section 3 introduces the proposed AWM method in detail. Section 4 presents the experimental results and discusses the contribution of performance improvement. We conclude our paper in Section 5.

\section{Related Works}
\begin{figure}
    \begin{center}
    \includegraphics[width=8.0cm]{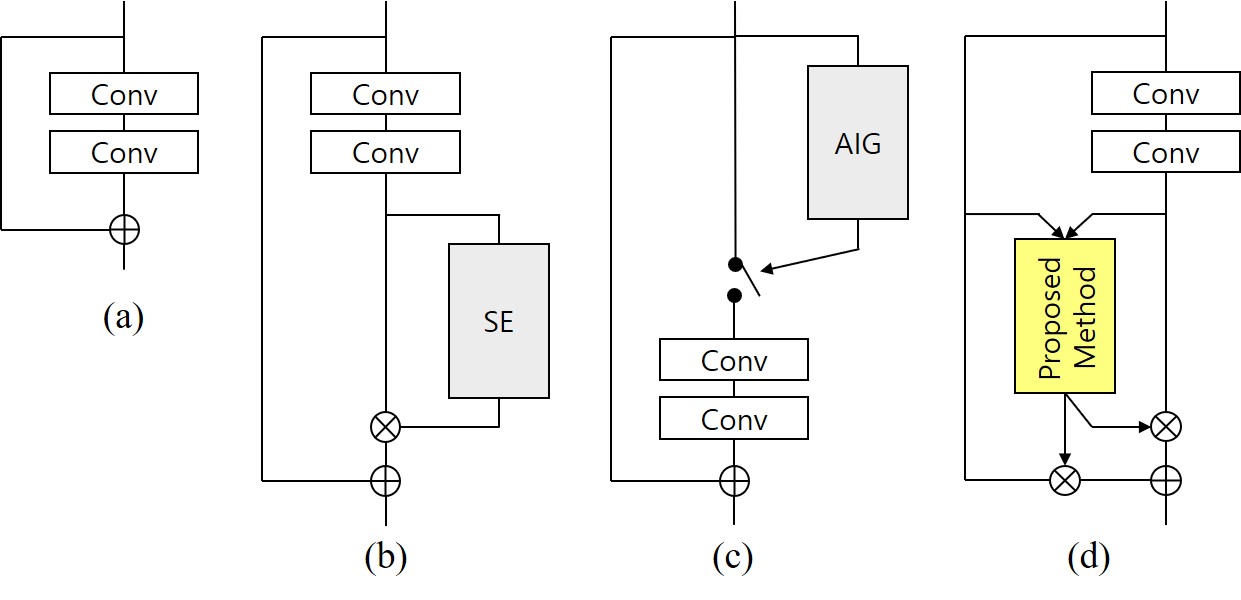}
    \end{center}
    \caption{Schematics of different network architectures for comparison. (a) ResNet~\cite{Res} for identical mapping, (b) SE module~\cite{SE} for modeling the channel relationship, (c) AIG module~\cite{AIG} for the gating mechanism, and (d) the proposed module for the active weighted mapping.}
    \label{fig:2}
\end{figure}

In terms of the network architectures during deep learning, after stacking over ten CNN layers of VGGNet\cite{VGG} and GoogLeNet~\cite{Goo}, ResNet~\cite{Res} has been one of the most successful approaches with an identity shortcut connection, as shown in Fig.~\ref{fig:2} (a). To achieve better performances, the stochastic depth-based network~\cite{Sto} improved the ResNet architecture by randomly dropping layers during training to allow better information and gradient flow. ResNeXt~\cite{Rxt} increased the cardinality of a network without involving many parameters. However, it was assumed that stacking with deep mapping by identical mapping was a good solution. This assumption may not be entirely valid, and thus, we attempted to improve the identical mapping by modifying it in the form of the proposed AWM method.

Our work is inspired from the squeeze-and-excitation module~\cite{SE}, as shown in Fig.~\ref{fig:2} (b), which explicitly models the interdependencies between the channel information; this module directly influences the design of the attention mechanism such as the convolutional block attention module~\cite{ReA}\cite{CBA}. In our work, the convolutional feature embedding method is used not to implement attention but to model the output values of each unit as the channel-wise extracted feature to determine which unit is important on the fly at each mapping unit. Another approach is the convolutional network with AIG~\cite{AIG}, as shown in Fig.~\ref{fig:2} (c), which adaptively define the network topology conditioned on the input image. For each layer, a proposed gate mechanism determines whether to execute or skip the layer; this method ensures a higher accuracy with low additional computational complexity. In this respect, this method represents hard attention while our proposed method involves soft weighted mapping. SkipNet~\cite{Ski} also use the gate-based mechanism for selecting the convolutional layers similar to AIG. The proposed method does not skip or discard any information in the current layer, even if it is not important, because the information might be used on the next upper layers for visual recognition. In this respect, there is a difference in the objectives of AIG and our proposed method. From the viewpoint of the network architecture, as shown in Fig.~\ref{fig:2} (d), we need two inputs from different paths in order to evaluate which of the two paths is important.

\section{Proposed Method}
The ResNet~\cite{Res} consists of $N$ layers whose basic components are residual convolutional blocks and the corresponding shortcut connections. The basic components are merged by element-wise summation at the point of the identical mapping units. The number of identical mapping units is presented according to the number of layers, $N$, in Table~\ref{table:1}.
In this paper, we have an interest to the revision of the mapping units where the number of the mapping units increase as the stacked layer gets deeper.

\begin{table}[t]
\caption{Number of identical mapping units according to the number of ResNet layers.}
\label{table:1} \centering
\begin{tabular}{c|c}
    \hline
    Num. of Layers ($N$) & Num. of Mapping Unit   \\
	\hline
	\hline
    14	&    6\\
	\hline
    20	&    9\\
	\hline
    32	&    15\\
	\hline
    44	&    21\\
	\hline
    56	&    27\\
    \hline
    110	&   54\\
	\hline
\end{tabular}
\end{table}

\subsection{Active Weighted Mapping Module}

\begin{figure*}
    \begin{center}
    \includegraphics[width=15.0cm]{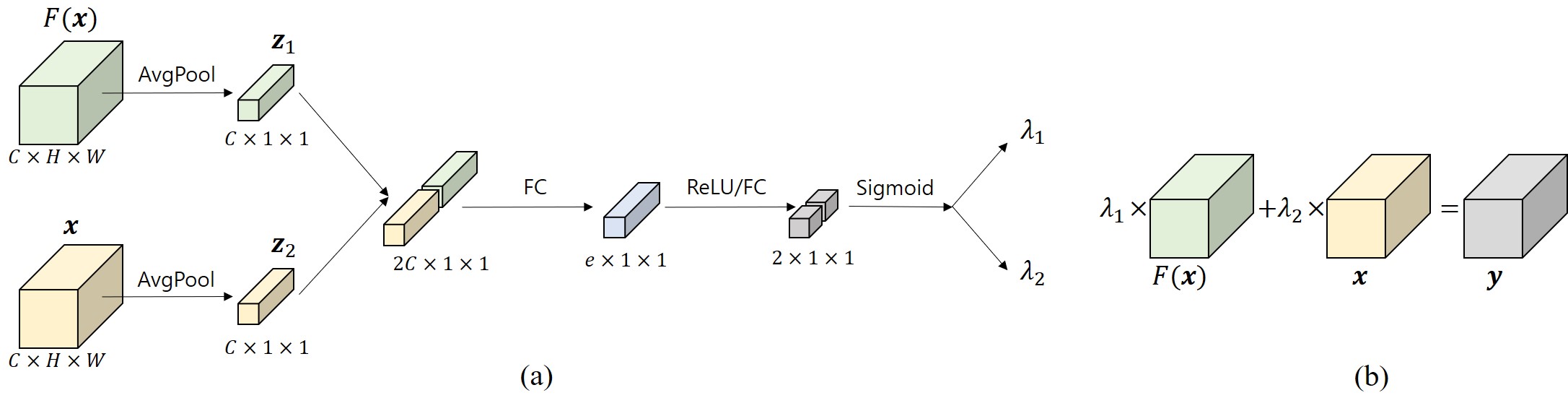}
    \end{center}
    \caption{Schematic of the proposed weighted mapping module. (a) Proposed active weighted mapping module, and (b) modified weight summation operation. FC denotes a fully connected layer. The weight values are normalized after the sigmoid layer. }
    \label{fig:3}
\end{figure*}

The basic identity mapping by shortcuts at~\cite{Res} is defined as:
\begin{equation}
    \mathbf{y}_k=F_k(\mathbf{x}_k)+\mathbf{x}_k
	\label{eq:1}
\end{equation}
where $\mathbf{x}_k$ and $\mathbf{y}_k$ are respectively the input and output vectors of the $k$th convolutional block $F_k$. Note that the dimensionality of $\mathbf{x}$ and $\mathbf{y}$ is $\Re^{(C\times H\times W)}$. $C$, $W$, and $H$ denote the channel size, width, and height of an input, respectively. In~\cite{Res}, it was assumed that identity mapping is sufficient for stacking more number of layers and it is economical; however, in this paper, we fundamentally redesign the identical mapping unit using the following weighted mapping module for efficient transfer of information according to the class of the input image. This approach can be written as
\begin{equation}
    \mathbf{y}_k=\lambda_{k1} F_k (\mathbf{x}_k )+\lambda_{k2} \mathbf{x}_k
    \label{eq:2}
\end{equation}
where $\lambda_{k1}$ and $\lambda_{k2}$ are the weight values. Specifically, we make use of two paths, i.e., shortcut connection and convolution units, simultaneously, unlike the other methods~\cite{SE}\cite{AIG} that pay attention to only a single path. We use the global averaged channel information as the representation of the two units to efficiently exploit the convolutional feature maps without incurring the high computational complexity present in~\cite{SE}\cite{CBA}. A more complex feature representation may further improve the recognition performance.

The channel-wise descriptor, $\mathbf{Z}$, was first proposed for exploiting inter-channel dependencies and the size of $\mathbf{Z}$ could be reduced, for example, $\Re^{(C\times H\times W)}\rightarrow \Re^{(C\times 1\times 1)}$. This could pose a problem in visual recognition because the exploited information is too small, and it is more effective to determine the importance of two paths using two weight values. The $c$-th element of $\mathbf{Z}=\{\mathbf{z}_1,\mathbf{z}_2,...,\mathbf{z}_c\}$ is calculated as in the following equation:
\begin{equation}
    \mathbf{z}_c=\frac{1}{(H\times W)} \sum_{i=1}^H\sum_{j=1}^W \mathbf{x}_{c,i,j}
    \label{eq:3}
\end{equation}
where it could be implemented by the global average pooling layer.

The embedded channel information comes from two paths. As shown in Fig.~\ref{fig:3}, $\mathbf{Z}_1$ and $\mathbf{Z}_2$ are obtained from the convolutional unit and the shortcut connection unit, respectively. For modeling the dependency of multiple paths, we form the concatenated feature by $\mathbf{Z}=\{\mathbf{Z}_1,\mathbf{Z}_2\}$ and the non-linear fully connected layers are passed sequentially.
\begin{equation}
    \lambda = \sigma (\mathbf{W}_2 \delta(\mathbf{W}_1 \mathbf{Z}))
    \label{eq:4}
\end{equation}
where $\sigma$ denotes the sigmoid function, the fully connected layers are $\mathbf{W}_1$, $\Re^{(e\times 2C)}$, and $\mathbf{W}_2$, $\Re^{(2\times e)}$, and $\delta$ is the rectified linear unit (ReLU) function for non-linearity. The reduced dimensionality, $e$, of the hidden layer is in the range $2<e<2C$. In the end, we have $\lambda=\{\lambda_1,\lambda_2\}$ and the weighted values are used after normalization.

\subsection{Training Strategy}

\begin{figure}
    \begin{center}
    \includegraphics[width=8.5cm]{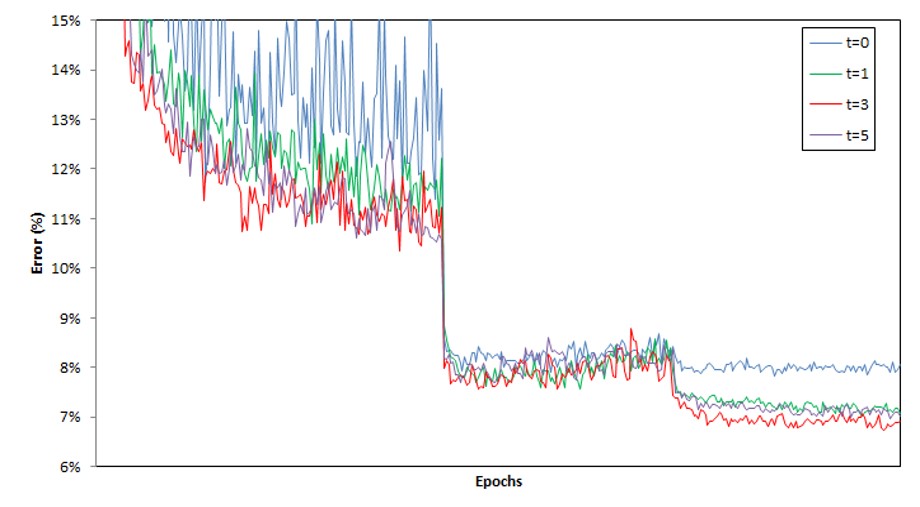}
    \end{center}
    \caption{Test error on Cifar-10 with ResNet using 20 layers. We learn each model under alternate $t$ epochs. }
    \label{fig:4}
\end{figure}


In the training stage, \cite{AIG} method assumed that only binary decisions were strictly considered, e.g., hard constraints such as "open" or "close" for selecting the layers, which is likely to cause the model collapse. On the other hand, our proposed method assume that a soft constraint-based network architecture using the inferred weight values without layer selections. However, it is not free from the mode collapse issues because the backbone network and the proposed AWM module are coupled together. For stable training of the proposed method, we train the backbone network architecture, e.g., ResNet, and the AWM module alternately. For example, in the early stage of learning, the parameters of the convolutional layer at the backbone architecture might be not properly learned for visual recognition, and the AWM modules also do not work correctly because they are derived from the backbone network.
To overcome this issue, we learn the backbone network and the AWM modules alternately. Specifically, when the backbone network is learning, the AWM modules are frozen, and vice versa. Note that when learning the backbone architecture for the first epoch, we set the weight values to 0.5.
In this case, the number of epochs to learn each model, $t$, is important and Fig.~\ref{fig:4} shows the test error changes according to various parameters, $t$, in the ResNet using 20 layers. In case of $t=0$, the backbone network and the proposed method are simultaneously learned as mentioned in~\cite{SE}\cite{AIG}; however, the corresponding performance in this case is worse than that for the others. As shown in Fig.~\ref{fig:4}, the alternative training strategy works successfully, and this straightforward approach could alleviate the model collapse during the training procedure. The best accuracy is achieved at $t=3$.

\section{Experimental Results and Discussion}
In this section, we describe the experiments performed to investigate the effectiveness of the proposed method across a range of datasets and model architectures. We also check the discriminant power of the inferred weight values in visual recognition.

\subsection{Experimental Results for Cifar-10 and Cifar-100 datasets}

\begin{table}[t]
\caption{Top-1 error rates on the Cifar-10 and Cifar-100 databases. ResNet and the proposed method are implemented using PyTorch.}
\label{table:2} \centering
\begin{tabular}{c|c|c|c|c}
    \hline
    Database & \multicolumn{2}{c|}{Cifar-10} & \multicolumn{2}{c}{Cifar-100} \\
    \hline
    Layers & ResNet & Proposed &  ResNet & Proposed \\
	\hline
	\hline
    14	&9.21\%	&8.13\%	&34.01\%	&31.77\%	\\
    \hline
    20	&8.42\%	&6.73\%		&32.62\%	&30.57\%	\\
    \hline
    32	&7.46\%	&6.09\%	&31.47\%	&28.88\%	\\
    \hline
    44	&7.16\%	&5.75\%	&29.62\%	&27.70\%	\\
    \hline
    56	&6.96\%	&5.54\%		&28.90\%	&27.09\%	\\
    \hline
    110	&6.61\%	&5.23\%	&27.81\%	&25.54\%	\\
	\hline
\end{tabular}
\end{table}

\textbf{Baseline Experiments }
To validate the proposed method, we first evaluate our method on the representative benchmark databases, i.e., Cifar-10 and Cifar-100~\cite{Cif}. Cifar-10 and Cifar-100 consist of 50,000 training images and 10,000 test images with $32\times32$-pixel color images. The only difference between Cifar-10 and Cifar-100 is the number of classes: 10 and 100 classes, respectively. We use the basic data augmentations for Cifar-10/100 databases, for example, zero padding with 4-pixels on each side, randomly cropped $32\times32$ pixel-based images, and random horizontal flipping. Our code and other baseline methods are implemented on the PyTorch and TorchVision framework~\cite{Tor}. In this paper, we train the proposed model using the stochastic gradient descent (SGD) with the Netstrov momentum for 350 epochs. The initial learning rate is set to 0.1, and it is decayed by a factor of 0.1 at 150 and 250 epochs. We use the following hyperparameters: a weight decay of 0.0001, momentum of 0.9, and batch size of 128. Moreover, the dimension reduction of the hidden layers, $e$, is set to 16 and the alternate switching parameter, $t$, is 3 in this experiment.

We use ResNet as the baseline network architecture for performance comparison and from Table~\ref{table:2}, we confirm that the performance is improved regardless of the increase in the number of layers, e.g., from 14 to 110 layers. Table~\ref{table:2} shows that the performance improvement is ensured not only in Cifar-10 but also in Cifar-100. Specifically, the average improvements in the top-1 error rates of Cifar-10 and Cifar-100 are 1.39\% and 2.15\%, respectively. From this result, we can conclude that the proposed AWM method works successfully compared with the previous identical mapping method on the ResNet of varying depth.

\begin{figure}
    \begin{center}
    \includegraphics[width=8.0cm]{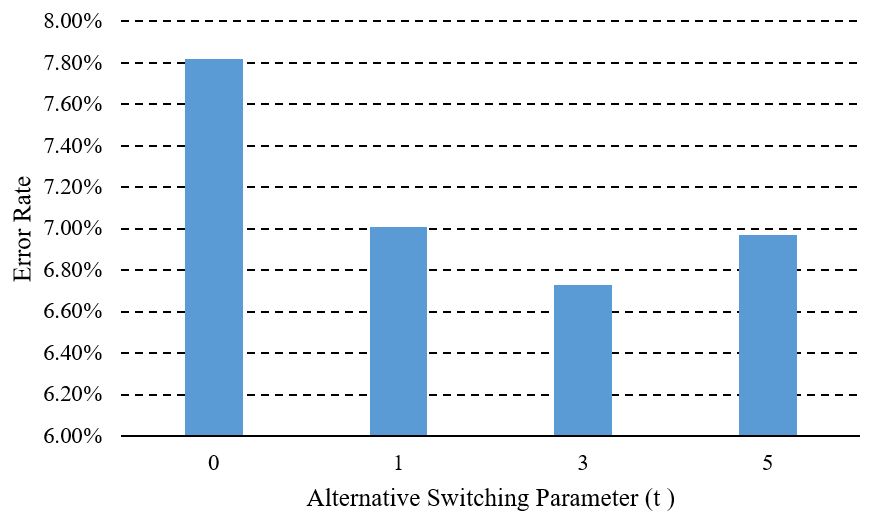}
    \end{center}
    \caption{Change in error rates with different alternate switching parameters using ResNet with 20 layers on Cifar-10. }
    \label{fig:5}
\end{figure}

We now consider the changes in accuracy with different values of alternate switching parameter, $t$, in Fig.~\ref{fig:5}. As described in Section 3.1, the best accuracy is achieved at $t=3$; the performance gap between $t=0$ and $t=3$ is 1.09\% and we can see that an excessive increase t leads to performance degradation. This is mainly because a large $t$ value leads to independent learning between the backbone network and the proposed mapping unit, which finally results in inferior performance. In this paper, we used $t=3$ for simplicity.

\begin{figure*}
    \begin{center}
    \includegraphics[width=17.5cm]{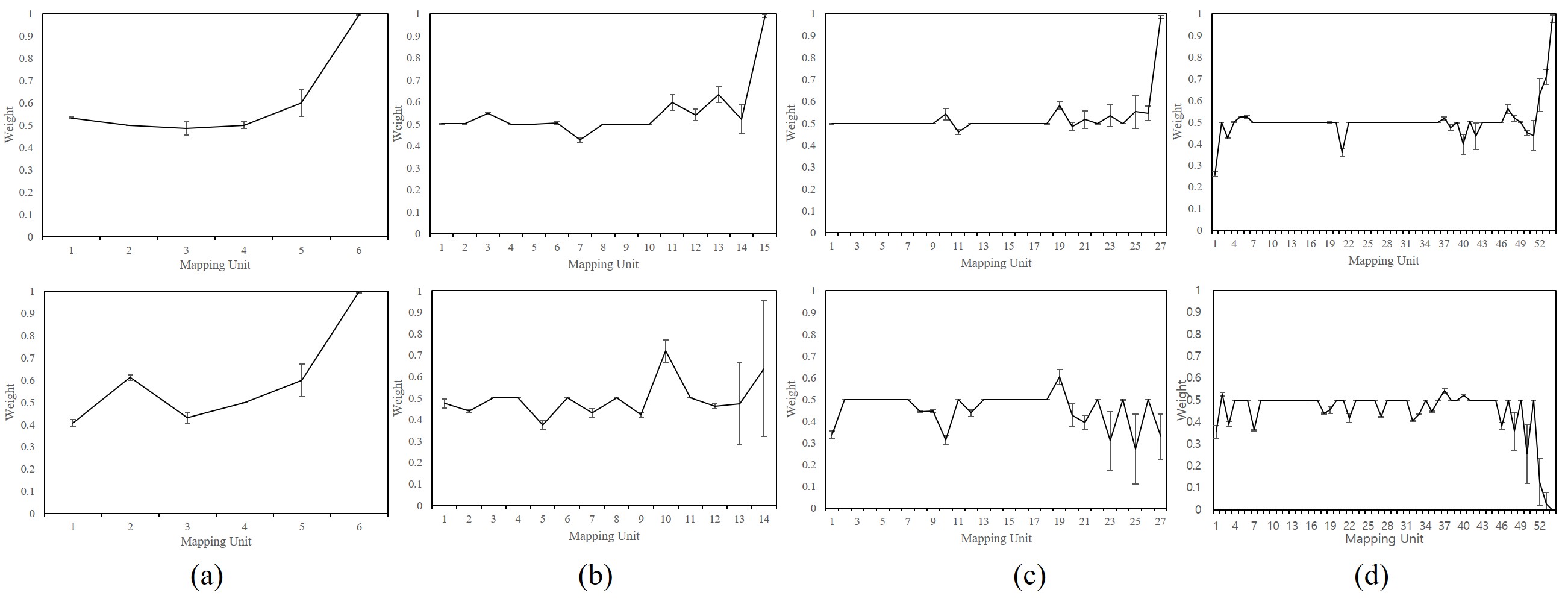}
    \end{center}
    \caption{Inferred weight parameters on Cifar-10 (top row) and Cifar-100 (bottom row). The conventional identical mapping is 0.5 through all mapping units; however, the proposed method uses different weights according to the class of the input image. The weight on the Y axis indicates the $\lambda_1$ value. The line and the bar indicate the average and the variance of the weight values, respectively. (a) 14-layer, (b) 32-layer, (c) 56-layer, and (d) 110-layer-based ResNets are used as the basic backbone network with the proposed method. }
    \label{fig:6}
\end{figure*}

\textbf{Analysis on inferred weights }
Fig.~\ref{fig:6} shows the summary of the different weight values inferred by the proposed method on the Cifar-10 and Cifar-100 datasets. As shown in Figs. 6 (a) and (b), the proposed method produces weight values that are biased towards the convolutional blocks rather than the shortcut connection, because the backbone network has only 14 layers. Note that when the network moves to a deeper configuration,
the number of the mapping units whose inferred weights are approximately 0.5 within small variances increases.
Because the proposed method learns the backbone network and the AWM module alternately, equal weights are preferred for stacking deep layers steadily at deep networks, and the weight values are biased to one side when different weights are really needed. As shown in Figs. 6 (c) and (d), if the backbone network is very deep, the first weight value of the active mapping module is biased to the shortcut connection rather than the convolutional blocks of the residual block. We assume that the first convolutional layer independent of the consecutive residual blocks plays a pivotal role in analyzing the basic edge information of an input image, and if possible, it is preferable that more edge information is delivered to the backward convolutional blocks for better accuracy.

It should also be noted that the weight values of the last residual block are different between Cifar-10 and Cifar-100. The weight is biased towards the convolutional blocks on Cifar-10 while the bias is towards the shortcut connection in Cifar-100. We believe that the network for Cifar-10 needs more convolutional blocks for further image analysis because compared to Cifar-100, Cifar-10 has smaller classes and larger inter-class variations. However, the variances of the weight values on Cifar-100 are larger, which means that the change in the weights is large to efficiently represent many classes. In this respect, we conclude that the proposed method could produce different weight values from the mapping units of the deep network, and their values are changed according to the characteristic of the input image. In Section 4.4, we describe the investigation concerning the discriminant powers of the weight values for visual recognition.

\begin{figure}
    \begin{center}
    \includegraphics[width=7.0cm]{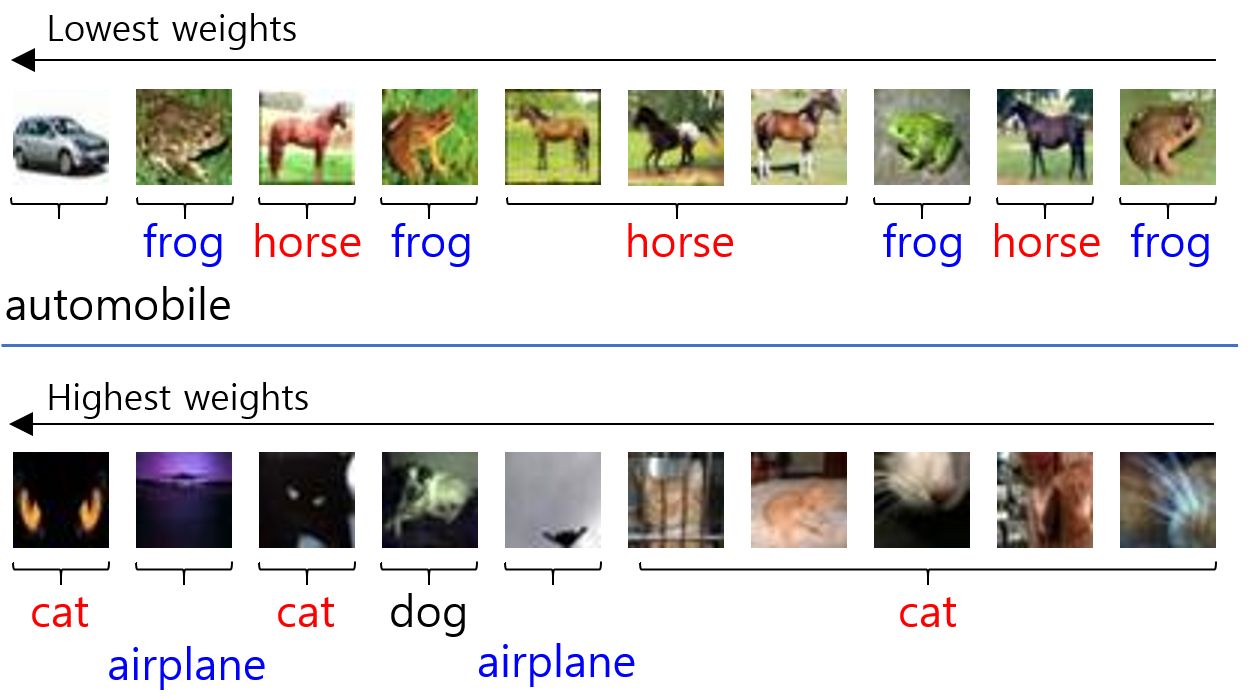}
    \end{center}
    \caption{Visualization of example images according to the summation of the weight values of the convolutional blocks. We use the ResNet with 110 layers in Cifar-10. Top images are sorted by lower weight values and bottom images are arranged in desending order of the weight values. } 
    \label{fig:vis}
\end{figure}

We investigate the example images for better understanding the inferred weighted values made by the proposed method. After simply summing the weight values, $\lambda_1$, of the convolutional blocks, we sort the images corresponding to the highest weight values and the lowest weight values like Fig.~\ref{fig:vis}. Note that the highest weight values of the convolutional blocks means that the network depends on more convolutional blocks for predicting the class of the input image. As shown in bottom of Fig.~\ref{fig:vis}, most images are a cat and an airplane. Note that these classes have the large image variations, for example, only a part of a cat is shown or an airplane is taken very small in an image. On the other hand, the top images of Fig.~\ref{fig:vis} are a frog and a horse, and the image variation of these classes is small. All in all, we can confirm that the weight values are similar if the class of the input images is the same.

\begin{table}[t]
\caption{Comparison with well-known studies performed on the Cifar-10 database. The backbone architecture employed is ResNet with 110 layers. }
\label{table:3} \centering
\begin{tabular}{c|c|c}
    \hline
    Method & Top-1 Error & Param \\
	\hline
	\hline
    ResNet~\cite{Res}	&6.43\%	&1.7M\\
    PreActivated ResNet~\cite{Res-1}	&6.37\%	&1.7M\\
    Stochastic Depth ResNet~\cite{Sto}	&5.25\%	&1.7M\\
    PyramidNet~\cite{Pyr}	&4.62\%	&1.7M\\
    DenseNet~\cite{Den}	&3.74\%	&27.2M\\
    ConvNet-AIG~\cite{AIG}	&5.76\%	&1.78M\\
    SkipNet~\cite{Ski} & 6.40\% & -\\
    Proposed Method	&5.23\%	&1.78M\\
	\hline
\end{tabular}
\end{table}

\textbf{Comparison with well-known methods }
Table~\ref{table:3} shows the further quantitative comparison among the well-known methods~\cite{Res}\cite{Res-1}\cite{Sto}\cite{AIG} and a brief summary of the accuracy on the Cifar-10 database. For fair comparison, each accuracy is as mentioned in the corresponding paper, and the basic depth of all the approaches is 110 layers. The best accuracy (3.74\%) is achieved by the DenseNet~\cite{Den}, but the number of parameters required for building the convolutional weights is more than 10 times that of the others. Among the methods using approximately 1.7M parameters, the Stochastic Depth ResNet exhibits an accuracy of 5.25\% while the proposed method demonstrates a slightly lower accuracy of 5.23\%, From the viewpoint of the network complexity, the ResNet-based methods~\cite{Res}\cite{Res-1}\cite{Sto} use 1.7M parameters for the convolutional weights while the ConvNet-AIG method~\cite{AIG} as well as the proposed method use 1.78M parameters. Therefore, the parameter overhead is not significant compared with the accuracy improvement.

\subsection{Experimental Results on ImageNet2012 dataset}

\begin{table}[t]
\caption{Top-1 and Top-5 error rates on ImageNet 2012. The backbone architecture employed is ResNet. ResNet and the proposed method are implemented using PyTorch. }
\label{table:4} \centering
\begin{tabular}{c|c|c}
    \hline
    Method & Top-1 & Top-5 \\
	\hline
	\hline
ResNet (18 layer)	&30.08\%	&10.78\%\\
Proposed Method (18 layer)	&29.36\%	&10.41\%\\
\hline
ResNet (50 layer)~\cite{Res} & 24.7\% & 7.8\%\\
ResNet (50 layer) & 24.52\% &7.50\%\\
Proposed Method (50 layer) & 24.04\% & 7.31\%\\
	\hline
\end{tabular}
\end{table}

For further validation of the proposed method under a large set of categories, we use the ImageNet database~\cite{Img}, which is a well-known dataset for the ImageNet large scale visual recognition challenge (ILSVRC). This dataset consists of one million training images and 50,000 validation images and the number of classes is 1,000. We adopt the basic data augmentation scheme with~\cite{Res} for training and use a single-crop evaluation with $224\times224$ pixel-based image during testing. The basic hyper-parameters are as follows: 256 mini-batch, momentum of 0.9, and weight decay of $10^{-4}$; the learning rate starts from 0.1 and drops every 30 epochs. We train the proposed method with 18 layers for 100 epochs and report the classification error on the validation set. The main purpose of this experiment was not to achieve the best accuracy in the ImageNet database but to validate the generality of the proposed method. Table~\ref{table:4} summarizes the comparison results between the baseline method, ResNet~\cite{Res}, and the proposed method. We can see that the proposed method outperforms the baseline method from the Top-1 to Top-5 errors.

\subsection{Extended Experimental Results: DenseNet on Cifar-10 dataset}

\begin{figure}
    \begin{center}
    \includegraphics[width=8.0cm]{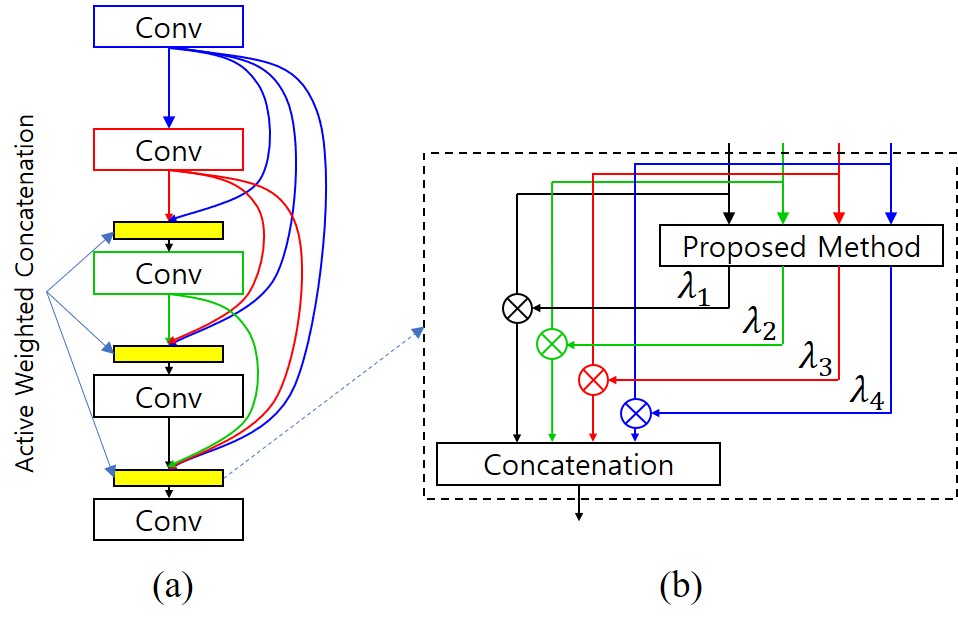}
    \end{center}
    \caption{Schematics of the (a) active weighted concatenation method for DenseNet and (b) the last proposed module. $\sum_i\lambda_i=1$}
    \label{fig:7}
\end{figure}

\begin{figure}
    \begin{center}
    \includegraphics[width=8.0cm]{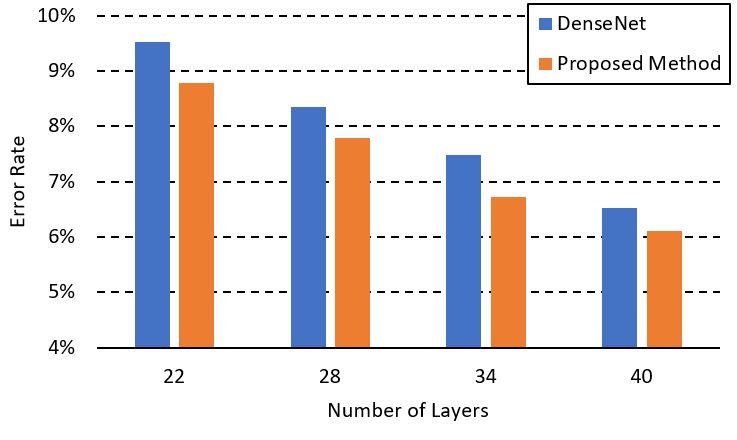}
    \end{center}
    \caption{Performance comparison through the different numbers of layers (e.g., from 22 layers to 40 layers) on Cifar-10. The used backbone network is DenseNet~\cite{Den}.}
    \label{fig:8}
\end{figure}

This experiment is designed to validate whether the proposed method demonstrates similar performance improvement in other backbone network such as DenseNet~\cite{Den}, in which each layer is connected to every other layer without using the identical mapping scheme. For improving the information flow between layers, all the connected layers are concatenated instead of summation~\cite{Res}. As shown in Fig.~\ref{fig:7}, we calculate the importance of each connection based on the proposed method, and before concatenation, we apply the different weight values to the corresponding connectivity. This is similar to the soft-assignment coding method~\cite{Sof}. For this experiment, we use the bottleneck-based DenseNet on Cifar-10 in which the augmentation scheme is used as in~\cite{Res} and the used hyper-parameters are the same used in~\cite{Den}. Fig.~\ref{fig:8} shows that the proposed method leads to better results, with an average improvement of +0.62\% in the error rate, compared with the DenseNet when the number of layers in changed from 22 to 40. These results indicate that the proposed method could be extended to the dense connectivity scheme, successfully.

\subsection{Experimental Results using LDA for Discriminant Power of Weights}

\begin{figure}
    \begin{center}
    \includegraphics[width=8.0cm]{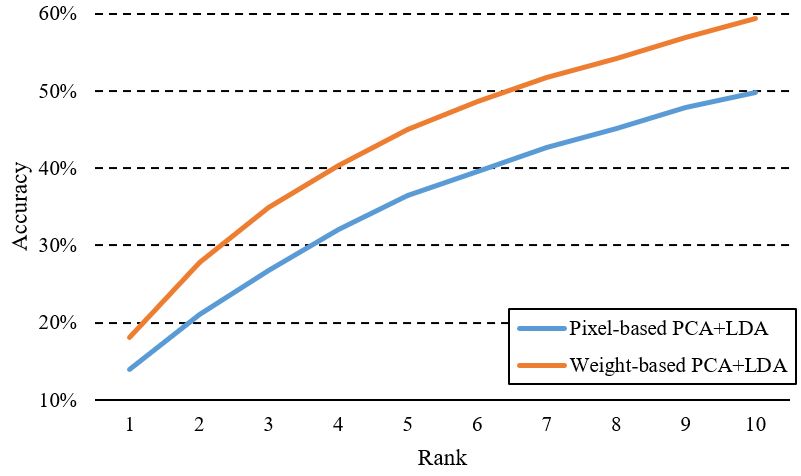}
    \end{center}
    \caption{ROC curves of two different feature-based PCA+LDA methods on Cifar-100. The weight value, $\Re^{54}$, inferred by the proposed method has higher discriminant power than the image, $\Re^{3,072}$, despite its low dimensionality.}
    \label{fig:9}
\end{figure}

For validating the discriminant power of the inferred weight values from each mapping module, we design the following ablation test using the LDA feature-based k-NN method~\cite{Dud} with the class averages. For building the baseline method, the image of Cifar-100 set is converted into a vector with a dimensionality of 3,072 ($=32\times32\times3$ pixels). We apply the principal component analysis (PCA) and LDA~\cite{LDA} to gain the discriminant power and reduce the feature dimensionality from $\Re^{3,072}$ to $\Re^{30}$. This is termed as the "pixel-based PCA+LDA." From the training set, each average feature, $m=\{1,2,3,…,100\}$ and $\Re^{30}$, of all the 100 classes is calculated. An extracted feature from the test set is compared with these 100 average features, m, using the Euclidean distance. We select the average feature with the shortest distance, and if the class of the selected average feature is identical to the true class of a test image, the Rank-1 accuracy ($k=1$) increases. In case of the proposed method, we use ResNet with 110 layers as the basic backbone network and extract the complete weight values from 54 mapping modules. Then, we have 50,000 vectors as the training set and the dimensionality of each vector is 54. We apply PCA and LDA, and the final dimension of the extracted feature is 30 like the baseline method, "pixel-based PCA+LDA." We term this "weight-based PCA+LDA." Fig.~\ref{fig:9} shows the ROC curves of the two PCA+LDA methods. The weight-based PCA+LDA outperforms the pixel-based PCA+LDA by an average of 8.14\% from Rank-1 to Rank-10. Thus, we observe that the weight values extracted from the proposed AWM modules have their own discriminant power compared with the image. Note that the feature dimension is only 54, and not 3,072. In this respect, we conclude that the weight values of the proposed method are different according to the class of an image, and thus, it is advantageous to stack more ResNet layers using the proposed AWM module.

\section{Conclusion}

In this paper, we introduced the AWM module designed to modify the identical mapping of the ResNet network, which is a well-known deep learning architecture in visual recognition. Using the proposed method, we obtain different weight values according to the different classes of images on the fly, and they are then used to efficiently merge the forward-feed information comes from both the shortcut connection and the convolutional blocks. We demonstrated the effectiveness and the generality of our approach through the results obtained by extensive experiments. In particular, the proposed method works well irrespective of backbone network employed, such as Resnet and DenseNet. To carry out quantitative analysis for the dependence of the weight values on the class, simple experiments were performed for the comparison of relative performance attained using the LDA method with the extracted weight values. Despite the success of ResNet, we believe that the ResNet style network architecture could be improved, and this works represents an attempt to do so.



{\small
\begin{thebibliography}{10}\itemsep=-1pt

\bibitem{Tor}
Pytorch.
\newblock {\em http://pytorch.org}.

\bibitem{Dud}
R.~O. Duda, P.~E. Hart, and D.~G. Stork.
\newblock Pattern classification.
\newblock {\em Wiley-Interscience, 2 edition}, 2000.

\bibitem{Pyr}
D.~Han, J.~Kim, and J.~Kim.
\newblock Deep pyramidal residual networks.
\newblock {\em IEEE Conf. on Computer Vision and Pattern Recognition}, pages
  6307--6315, 2017.

\bibitem{Res}
K.~He, X.~Zhang, S.~Ren, and J.~Sun.
\newblock Deep residual learning for image recognition.
\newblock {\em IEEE Conf. on Computer Vision and Pattern Recognition}, pages
  770--778, 2016.

\bibitem{Res-1}
K.~He, X.~Zhang, S.~Ren, and J.~Sun.
\newblock Identity mappings in deep residual networks.
\newblock {\em European Conf. on Computer Vision}, pages 630--645, 2016.

\bibitem{SE}
J.~Hu, L.~Shen, and G.~Sun.
\newblock Squeeze-and-excitation networks.
\newblock {\em IEEE Conf. on Computer Vision and Pattern Recognition}, Jun.
  2018.

\bibitem{Den}
G.~Huang, Z.~Liu, L.~van~der Maaten, and K.~Weinberger.
\newblock Densely connected convolutional networks.
\newblock {\em IEEE Conf. on Computer Vision and Pattern Recognition}, pages
  2261--2269, 2017.

\bibitem{Sto}
G.~Huang, Y.~Sun, Z.~Liu, D.~Sedra, and K.~Q. Weinberger.
\newblock Deep networks with stochastic depth.
\newblock {\em European Conf. on Computer Vision}, 4:646--661, 2016.

\bibitem{Ale}
A.~Krizhevsk, I.~Sulskever, and G.~E. Hinton.
\newblock Imagenet classification with deep convolutional neural networks.
\newblock {\em Neural Information Processing Systems}, 25:1097--1105, 2012.

\bibitem{Cif}
A.~Krizhevsky.
\newblock Learning multiple layers of feature from tiny images.
\newblock {\em Tech report}, 2009.

\bibitem{Sof}
L.~Liu, L.~Wang, and X.~Liu.
\newblock In defense of soft-assignment coding.
\newblock {\em IEEE Conf. on Computer Vision}, pages 2486--2493, 2011.

\bibitem{LDA}
A.~Martinez and A.~Kak.
\newblock Pca versus lda.
\newblock {\em IEEE Trans. Pattern Recognition and Machine Intelligence},
  23(2):228--233, Feb. 2001.

\bibitem{Img}
O.~Russakovsky, J.~Deng, H.~Su, J.~Krause, S.~Satheesh, S.~Ma, Z.~Huang,
  A.~Karpathy, A.~Khosla, M.~Bernstein, A.~C. Berg, and L.~Fei-Fei.
\newblock Imagenet large scale visual recognition challenge.
\newblock {\em International Journal of Computer Vision}, 115(3):211--252,
  2015.

\bibitem{VGG}
K.~Simonyan and A.~Zisserman.
\newblock Very deep convolutional networks for large-scale image recognition.
\newblock {\em International Conference on Learning Representations}, 2015.

\bibitem{Hig}
R.~K. Srivastava, K.~Greff, and J.~Schmidhuber.
\newblock Highway networks.
\newblock {\em arXiv preprint}, page arXiv:1505.00387, 2015.

\bibitem{Goo}
C.~Szegedy, W.~Liua, Y.~Jia, P.~Sermanet, S.~Reed, D.~Anguelov, D.~Erhan,
  V.~Vanhoucke, and A.~Rabinovich.
\newblock Going deeper with convolutions.
\newblock {\em IEEE Conf. on Computer Vision and Pattern Recognition}, pages
  1--9, 2015.

\bibitem{AIG}
A.~Veit and S.~Belongie.
\newblock Convolutional networks with adaptive inference graphs.
\newblock {\em European Conf. on Computer Vision}, Sept. 2018.

\bibitem{Ens}
A.~Veit, M.~Wilber, and S.~Belongie.
\newblock Residual networks behave like ensembles of relatively shallow
  networks.
\newblock {\em Neural Information Processing Systems}, 29:550--558, 2016.

\bibitem{ReA}
F.~Wang, M.~Jiang, C.~Qian, S.~Yang, C.~Li, H.~Zhang, X.~Wang, and X.~Tang.
\newblock Residual attention network for image classification.
\newblock {\em arXiv preprint}, arXiv:1704.06904, 2017.

\bibitem{Ski}
X.~Wang, F.~Yu, Z.-Y. Dou, T.~Darrell, and J.~E. Gonzalez.
\newblock Skipnet: Learning dynamic routing in convolutional networks.
\newblock {\em European Conf. on Computer Vision}, Sept. 2018.

\bibitem{CBA}
S.~Woo, J.~Park, J.~Lee, and I.~Kweon.
\newblock Cbam: Convolutional block attention module.
\newblock {\em European Conf. on Computer Vision}, Sept. 2018.

\bibitem{Rxt}
S.~Xie, R.~Girshick, P.~Dollar, Z.~Tu, and K.~He.
\newblock Aggregated residual transformations for deep neural networks,.
\newblock {\em IEEE Conf. on Computer Vision and Pattern Recognition}, Jul.
  2017.

\bibitem{Wid}
S.~Zagoruyko and N.~Komodakis.
\newblock Wide residual network.
\newblock {\em British Machine Vision Conference}, pages 87.1--87.12, Sept.
  2016.

\end{thebibliography}

}
\end{document}